%% file: lrec-coling2024.tex
\documentclass[10pt, a4paper]{article}

\usepackage{lrec-coling2024} 

\usepackage{multirow}
\usepackage{graphicx}
\usepackage{arydshln}  
\usepackage{caption}
\usepackage{subcaption}
\usepackage{xurl}
\usepackage{amsmath}
\DeclareMathOperator*{\argmax}{argmax}

\title{Forget NLI, Use a Dictionary: Zero-Shot Topic Classification for Low-Resource Languages with Application to Luxembourgish}

\name{Fred Philippy\textsuperscript{1,2}, Shohreh Haddadan\textsuperscript{1}, Siwen Guo\textsuperscript{1}\vspace{0.2cm}} 

\address{\textsuperscript{\textbf{1}}Zortify S.A., Luxembourg\hspace{0.5cm}
         \textsuperscript{\textbf{2}}University of Luxembourg, Luxembourg \vspace{0.1cm}\\ \{\texttt{fred}, \texttt{siwen}\}\texttt{@zortify.com}, \texttt{shohreh.haddadan@gmail.com}\vspace{0.2cm}}

\input{00-abstract}

\begin{document}

\maketitleabstract

\input{00-abstract}
\input{01-intro}
\input{02-motivation}
\input{03-related}

\input{04-dataset}
\input{05-implementation}
\input{06-results}
\input{07-discussion}

\input{08-conclusion}

\input{limitations}
\input{ethics}

\section{Bibliographical References}\label{sec:reference}

\bibliographystyle{lrec-coling2024-natbib}
\bibliography{zotero}

\section{Language Resource References}
\label{lr:ref}
\bibliographystylelanguageresource{lrec-coling2024-natbib}
\bibliographylanguageresource{languageresource}
\appendix
\input{appendix}

\end{document}

%% file: 00-abstract.tex
\abstract{
In NLP, zero-shot classification (ZSC) is the task of assigning labels to textual data without any labeled examples for the target classes. A common method for ZSC is to fine-tune a language model on a Natural Language Inference (NLI) dataset and then use it to infer the entailment between the input document and the target labels. However, this approach faces certain challenges, particularly for languages with limited resources. In this paper, we propose an alternative solution that leverages dictionaries as a source of data for ZSC. We focus on Luxembourgish, a low-resource language spoken in Luxembourg, and construct two new topic relevance classification datasets based on a dictionary that provides various synonyms, word translations and example sentences. We evaluate the usability of our dataset and compare it with the NLI-based approach on two topic classification tasks in a zero-shot manner. Our results show that by using the dictionary-based dataset, the trained models outperform the ones following the NLI-based approach for ZSC. While we focus on a single low-resource language in this study, we believe that the efficacy of our approach can also transfer to other languages where such a dictionary is available.
 \\ \newline \Keywords{Less-Resourced/Endangered Languages, Document Classification, Corpus} }

%% file: 01-intro.tex
\section{Introduction}
Zero-shot classification (ZSC) allows to classify a text document into a category for which no labeled examples are available. A common technique for ZSC is to leverage pre-trained language models that have learned general semantic representations from large corpora. These models can be fine-tuned on a natural language inference (NLI) dataset and then be used to infer the entailment between the document and the labels \citep{yin_benchmarking_2019}. In this approach, each potential target label is considered as a hypothesis in natural language, and the NLI model is used to evaluate the level of entailment between the input document and potential labels. For example, given a document "I always eat my soup with a spoon" and the labels "food" and "animals", the model can predict a score of how likely the document entails each label. The label with the highest entailment score can be selected as the predicted class.

Directly adopting NLI datasets for ZSC poses several challenges and limitations in real-world scenarios. We identify and highlight three main limitations of such an approach. First, there is a mismatch between the NLI and ZSC tasks. Second, the performance of this approach depends on the availability and quality of NLI datasets, which are challenging and costly to obtain. Third, for many low-resource languages, the lack of pre-training data hinders the model's ability to solve complex reasoning tasks such as NLI. In this work, we discuss the case of Luxembourgish, a West Germanic language spoken by around 400,000 people in Luxembourg. There is no large NLI dataset for the language, and only a small amount of unlabeled pre-training data is available. Therefore, using NLI datasets for ZSC in Luxembourgish results in poor performance.

In this work, we propose an alternative solution that provides sufficient data for low-resource languages in the context of ZSC. The proposed approach exploits dictionaries as a source of data for ZSC. More specifically, this dictionary-based approach offers two main advantages: 1) it provides data that is more relevant to the task of ZSC, and 2) it leverages resources that are more readily available in many low-resource languages. We demonstrate our approach on the Luxembourgish language, for which we construct two new topic relevance classification datasets based on a dictionary.\footnote{Our code and datasets are accessible via \\ \url{https://github.com/fredxlpy/LETZ/}} In short, our main contributions are as follows:
\begin{enumerate}
    \setlength\itemsep{-0.1em}
    \item We introduce a new approach for creating datasets that allow to adapt models to ZSC for low-resource languages where a dictionary is available.
    \item Using this approach, we construct and release two new datasets for Luxembourgish that are more suitable for ZSC tasks than existing NLI datasets.
    \item We evaluate our datasets on the task of zero-shot topic classification by comparing the performance of models trained on our datasets and NLI datasets
\end{enumerate}

%% file: 02-motivation.tex
\section{Motivation}
Our work aims to address the following limitations and challenges that hinder the effectiveness of zero-shot classification for low-resource languages such as Luxembourgish:

\begin{enumerate}
\setlength\itemsep{-0.1em}
\item The mismatch between the fine-tuning task, NLI, and the inference task, topic classification, as the former requires reasoning about logical relations between sentences (entailment, contradiction, neutral), while the latter evaluates the relevance of labels to a sentence (relevant, irrelevant) \citep{ma_issues_2021}.

\item The difficulty and the expense of creating NLI data, especially for low-resource languages. NLI data requires high-quality annotations that capture the subtle nuances of entailment and contradiction between sentence pairs. Moreover, such annotations are often prone to inter-annotator disagreement, which undermines the validity and reliability of NLI datasets \citep{pavlick_inherent_2019, kalouli_curing_2023}.

\item The poor performance of language models on high-level tasks such as NLI for low-resource languages \citep{ebrahimi_americasnli_2022}. Low-resource language models suffer from insufficient training data and vocabulary coverage, which affects their ability to encode rich semantic representations and handle complex reasoning tasks such as NLI. \end{enumerate}

%% file: 03-related.tex
\section{Related work}

A common method for ZSC is the \textit{entailment approach} \citep{yin_benchmarking_2019}, which uses NLI datasets to fine-tune pre-trained language models and then apply them to ZSC tasks. However, this approach has several drawbacks, as discussed by \citet{ma_issues_2021}. They identify issues such as label mismatch, data imbalance, and semantic ambiguity that affect the performance and generalization of the entailment approach. Moreover, \citet{ebrahimi_americasnli_2022} show that NLI models perform cross-lingual transfer poorly for low-resource languages, which in turn affects their ZSC capability. Therefore, they argue for the need of creating annotated datasets for semantic tasks in low-resource languages.

\paragraph{Luxembourgish Language} \ \\
Luxembourgish is one of the three national languages of Luxembourg and is spoken by roughly 400,000 people ($\approx$ 70\% of the population). According to UNESCO \textit{World Atlas of Languages}\footnote{\url{https://en.wal.unesco.org}}, Luxembourgish belongs to the world's \textit{potentially vulnerable} languages.

However, Luxembourgish has seen significant transformations over the past century, including its development into a national language, expansion into written and digital media, and its role as a symbol of national identity. 

The sociolinguistic landscape of Luxembourg, with its unique multilingual setup \citep{purschke_sociolinguistics_2023} and the dynamic evolution of Luxembourgish from a dialect to a national language with increasing digital presence, provides a fertile ground for NLP research. Researching Luxembourgish through the lens of NLP contributes to the field of lesser-studied languages by developing methodologies that can be applied to other multilingual and language variation contexts.

%% file: 04-dataset.tex
\section{Our Dataset}
Based on a publicly available online dictionary, we create two new topic relevance classification datasets that allow to adapt pre-trained language models to zero-shot topic classification in Luxembourgish.

\subsection{Data Collection}
\textit{Luxembourg Online Dictionary}\footnote{\url{https://lod.lu}} (LOD) is a publicly available platform hosting a multilingual dictionary with the aim of promoting Luxembourgish as the language of communication, integration and literature. In the following, we present some statistics relevant to our work about the data provided by the Center for the Luxembourgish Language (ZLS\footnote{\textit{Zenter fir d'Lëtzebuerger Sprooch}}) in a report\footnote{\url{https://gouvernement.lu/fr/actualites/toutes_actualites/communiques/2022/06-juin/21-lod-neie-look.html}} in 2022.

The dictionary contains around \textbf{10,000 synonyms} and \textbf{48,000 example sentences} on approximately \textbf{31,000 entries}.  Words with multiple meanings are treated separately for each of their distinct meanings, with corresponding synonyms and example sentences. For most entries, the dictionary provides translations from/to 5 languages: German, French, English, Portuguese and Sign Language. In addition, it features 20,000 phonetic transcriptions, 30,000 audio recordings, 9,300 conjugation and declension tables as well as 5,000 proverbs and idiom explanations.

ZLS released all of their data on the Luxembourgish Open Data platform\footnote{\url{https://data.public.lu/en/organizations/zenter-fir-dletzebuerger-sprooch/}} under a \textit{Creative Commons Zero} (CC0) license. In this work, we use the dataset version released on June 5, 2023.

\subsection{From Dictionary to Dataset}

We first extract the part-of-speech tag, synonyms, and example sentences for each meaning of every word in the raw LOD data, and filter out the non-nouns.

Next, we assign all the synonyms of a word meaning as labels to its example sentences. To prevent the model from exploiting the shortcut of matching the label with the word occurrence in the sentence, we exclude the word itself from the label set .

Moreover, since many Luxembourgish words are orthographic variants of French or German words\footnote{ \underline{Examples}: “alerte” → “Alert”, “Million” → “Millioun”.}, we discard noun-synonym pairs that have a low Levenshtein distance.

Finally, we generate “non-entailment” samples by randomly selecting a word from the entire noun vocabulary as a label for each example sentence. However, we exclude any words that are similar to any of the words in the sentence based on the Levenshtein distance.

Following the exact same approach, we additionally create a separate dataset based on the word translations available in the dictionary instead of synonyms.

This new type of dataset is termed \textit{\textbf{L}uxembourgish \textbf{E}ntailment-based \textbf{T}opic classification via \textbf{Z}ero-shot learning} (LETZ), with the synonym-based dataset being referred to as \textbf{\texttt{LETZ-SYN}} and the one derived from word translations as \textbf{\texttt{LETZ-WoT}}.

The number of "entailment"/"relevant" ("1") and "non-entailment"/"irrelevant" ("0") samples is balanced for all sets. The dataset split sizes are provided in Table \ref{tab:ours_set_split}. We provide examples and more details of our data sets in Appendix \ref{app: our_dataset}.

\begin{table}[h]
    \centering
    \small
    \renewcommand{\arraystretch}{1.5}
    \begin{tabular}{c|c|c|c}
         Dataset & |Train| & |Dev| & |Test| \\ \hline
         \texttt{\texttt{LETZ-SYN}} &  11,822 & 1,478 & 1,478 \\
         \texttt{\texttt{LETZ-WoT}} & 39,132 & 4,892 & 4,892 \\ \hline
    \end{tabular}
    \caption{Dataset statistics.}
    \label{tab:ours_set_split}
\end{table}

\input{Tables/acc_and_f1}

%% file: Tables/acc_and_f1.tex
\begin{table*}[h!]
\centering
\renewcommand{\arraystretch}{1.5}
\begin{tabular}{c|c|cc|cc}
\hline
 &  & \multicolumn{2}{c|}{n = 568} & \multicolumn{2}{c}{n = 11.822} \\ \hline
Model & Train data & SIB-200 & LuxNews & SIB-200 & LuxNews \\ \hline
\multirow{6}{*}{mBERT} & NLI-lb & 17.52 (16.56) & 15.87 (12.51) & \textbackslash{} & \textbackslash{} \\
 & NLI-de & 25.61 (24.69) & 30.22 (25.88) & 48.04 (43.76) & 43.06 (35.18) \\
 & NLI-en & 22.67 (22.38) & 28.55 (23.20) & 49.51 (44.34) & 50.73 (38.18) \\
 & NLI-fr & 22.30 (21.30) & 25.02 (20.01) & 49.75 (45.77) & 46.30 (37.65) \\
 & \multicolumn{1}{l|}{LETZ-WoT} & \multicolumn{1}{l}{49.39 (49.50)} & \multicolumn{1}{l|}{59.81 (43.18)} & \multicolumn{1}{l}{53.55 (52.46)} & \multicolumn{1}{l}{59.96 (52.13)} \\
 & LETZ-SYN & \textbf{52.08 (51.45)} & \textbf{65.08 (49.20)} & \textbf{53.80 (54.13)} & \textbf{66.07 (47.73)} \\ \hline
\multirow{2}{*}{LuxemBERT} & NLI-lb & 14.58 (12.91) & 24.69 (16.53) & \textbackslash{} & \textbackslash{} \\
 & LETZ-SYN & \textbf{18.50 (15.86)} & \textbf{30.63 (19.48)} & \textbf{65.07 (64.07)} & \textbf{51.81 (38.27)} \\ \hline
\end{tabular}
\caption{Results of our experiments on two topic classification datasets. Experiments are conducted for different number of training samples \textbf{n} from the different training sets. The performance metrics are reported as "\textbf{Accuracy (F1 score)}" for each task.}
\label{tab:results}
\end{table*}


%% file: 05-implementation.tex
\section{Implementation}

\subsection{Training}
We conduct experiments using two different models that have been pre-trained on Luxembourgish data: \textbf{LuxemBERT} \citep{lothritz_luxembert_2022}, a monolingual Luxembourgish model, and \textbf{mBERT} \citep{devlin_bert_2019}, a multilingual BERT model that has been pre-trained on 102 languages, including Luxembourgish.

In order to perform the classification task, we append an additional layer to the pre-trained model that consists of a linear layer and a tanh activation function. The classification layer has two output nodes which are used to determine whether a given document contains a topic or not (Figure \ref{fig:fine-tuning}). Considering the limited amount of fine-tuning data, which could lead to variability in performance outcomes, we conduct each experiment four times using distinct random seeds. We then report the average results to account for any inconsistencies.

Besides fine-tuning both models on our new datasets, we use additional training datasets for comparison:
\begin{itemize}
    \item \textbf{NLI-lb} \citeplanguageresource{lothritz_luxembert_2022_lr}, a Luxembourgish NLI dataset consisting of 568 train and 63 validation samples. The dataset only contains entailment ("1") and contradiction samples ("0").
    \item \textbf{XNLI-de}, \textbf{XNLI-en} \& \textbf{XNLI-fr}, German, English and French subsets of the XNLI \citeplanguageresource{conneau_xnli_2018_lr} dataset respectively.
\end{itemize}

In addition, we perform experiments in "high-resource" (11,822 train and 1,478 validation samples)\footnote{Number of samples in \texttt{LETZ-SYN}.} and "low-resource" (568 train and 63 validation samples)\footnote{Number of samples in the Luxembourgish NLI dataset \citep{lothritz_luxembert_2022}.} settings.

\subsection{Evaluation}
Due to the inherent limitations associated with Luxembourgish being a low-resource language, there is a conspicuous lack of labeled datasets available. Within the context of topic classification, we could only identify two evaluation datasets that were suitable for our study:

\begin{itemize}
    \item The Luxembourgish subset of \textbf{\textit{SIB-200}} \citeplanguageresource{adelani_sib-200_2024}, a multilingual topic classification dataset, containing seven categories, namely: \texttt{science/technology}, \texttt{travel}, \texttt{politics}, \texttt{sports}, \texttt{health}, \texttt{entertainment}, and \texttt{geography}.
    \item A Luxembourgish News Classification dataset introduced by \citetlanguageresource{lothritz_luxembert_2022_lr}, consisting of news articles from a Luxembourg-based news platform. For our experiments we restrict it to the following 5 (out of 8) categories: \texttt{Sports}, \texttt{Culture}, \texttt{Gaming}, \texttt{Technology}, \texttt{Cooking recipes}. We exclude \texttt{National news}, \texttt{International news} and \texttt{European news} to avoid overlap with other categories. In what follows we will refer to this dataset as \textbf{\textit{LuxNews}}.

\end{itemize}

Following \citet{yin_benchmarking_2019}, we use an entailment approach (Figure \ref{fig:evaluation} in Appendix \ref{app: implementation}) to evaluate the models on these datasets, instead of a traditional supervised classification approach, where the number of output nodes corresponds to the number of categories. To be more exact, for a given sample $\textbf{x}$ and potential topics/categories $T = \{T_1, \ldots, T_n \}$, we compute the entailment probability for each pair $(\textbf{x}, T_i)_{i \in \{1,\ldots,n\}}$ denoted by $\textbf{P}_{i,1}$ and select $T_{i^*}$ where

$$
i^* = \argmax_{i \in \{1,\ldots,n\}} \textbf{P}_{i,1}
$$

The details of the training and evaluation methodology and the datasets employed are presented in Appendix \ref{app: implementation}.

%% file: 06-results.tex
\section{Results}
Table \ref{tab:results} shows that models fine-tuned on our datasets exceed the performance of those trained on NLI data, especially in the "low-resource" setting. More exactly, mBERT, with only 568 samples from our dictionary-based datasets, exceeds the results achieved with 20x more NLI samples in French, German, or English.

However, fine-tuning on German, French, or English NLI datasets markedly improves results over Luxembourgish data for which the performance is comparable to that of the random baseline. This suggests that the limited size of the Luxembourgish pre-training corpus may hinder the model’s ability to acquire a sufficient level of semantic and pragmatic understanding to solve complex reasoning tasks such as NLI.

 In the "low-resource" setting, LuxemBERT underperforms mBERT, suggesting it needs more data for task-specific knowledge compared to mBERT's general cross-lingual knowledge acquired during pre-training from high-resource languages. Nonetheless, in the "high-resource" setting, LuxemBERT outperforms mBERT on \textit{SIB-200} but underperforms on \textit{LuxNews}, possibly due to its inability to interpret multilingual speech excerpts or quotes.

%% file: 07-discussion.tex
\section{Discussion}
While we focus on Luxembourgish as an example of low-resource languages in this paper, we believe that this approach can be generalized to other languages where such dictionaries are available as well.

While we acknowledge that our method depends on the availability of dictionaries for low-resource languages, it is crucial to note that dictionaries often receive priority due to their fundamental role in educational and cultural preservation efforts. They are typically more prevalent because they form the bedrock for literacy and basic education, which are more fundamental needs than specialized datasets like those required for NLI. The creation of NLI datasets demands advanced linguistic knowledge and resources, making it a less immediate concern compared to building basic language tools. Initiatives, such as the \textit{Dictionaria}\footnote{\url{https://dictionaria.clld.org}} journal, the \textit{Living Dictionaries}\footnote{\url{https://livingdictionaries.app}} or the \textit{Webonary}\footnote{\url{https://www.webonary.org}} platform, support the development of dictionaries for low-resource and even indigenous languages. So, while both dictionaries and NLI datasets may not be universally available, there is a stronger, more widespread motivation behind the creation of dictionaries, rendering them relatively more accessible and likely to exist for low-resource languages.

Additionally, our experiments suggest that these dictionaries would not require tens of thousand of entries to be effective, as it appears that a multilingual language model can attain satisfactory performance with just a few hundred sentence-synonym or sentence-word translation pairs.

%% file: 08-conclusion.tex
\section{Conclusion}
This paper presents a new but simple approach to construct datasets that enable a language model to perform zero-shot topic classification in a low-resource language, such as Luxembourgish. We argue that the conventional approach of transferring from NLI to ZSC is ineffective for such languages, due to the semantic complexity of NLI and the scarcity of linguistic resources. We propose an alternative approach that leverages a dictionary to create a dataset that is more aligned with the ZSC task. We demonstrate that our dataset enables the model to outperform the ones that employ cross-lingual NLI transfer or in-language NLI fine-tuning on Luxembourgish ZSC, using over 20 times fewer training samples. In future work, we intend to explore the effectiveness of our approach when applied to other low-resource languages, as well as to high-resource ones.

%% file: limitations.tex
\section*{Limitations}
One of the limitations of our study is that we only focus on a single low-resource language, Luxembourgish, and we do not test our approach on other languages. Therefore, the generalizability of our method may be limited by the availability and quality of dictionaries for different languages. Another limitation is that we rely on a single source of data, namely a dictionary, which may not capture all the nuances and variations of natural language.

%% file: ethics.tex
\section*{Ethics Statement}
Our study aims to provide a novel solution for zero-shot classification in low-resource languages, which can potentially benefit various applications and users who need to classify textual data without labeled examples. 
While our method could potentially benefit any language, we specifically emphasize its usefulness for low-resource languages that suffer from data scarcity and lack of adequate tools. We believe that our method can contribute to the promotion of linguistic diversity, as well as to the empowerment and inclusion of speakers of low-resource languages.

However, we also acknowledge that some dictionaries may contain outdated, inaccurate, or offensive information that could harm certain groups or individuals. Therefore, we urge future researchers and practitioners to carefully select and evaluate the dictionaries they use and to adhere to the ethical principles and guidelines of their respective fields and communities.

%% file: appendix.tex
\section{Our Dataset} \label{app: our_dataset}
\begin{figure}[!ht]
    \centering
    \includegraphics[width=0.48\textwidth]{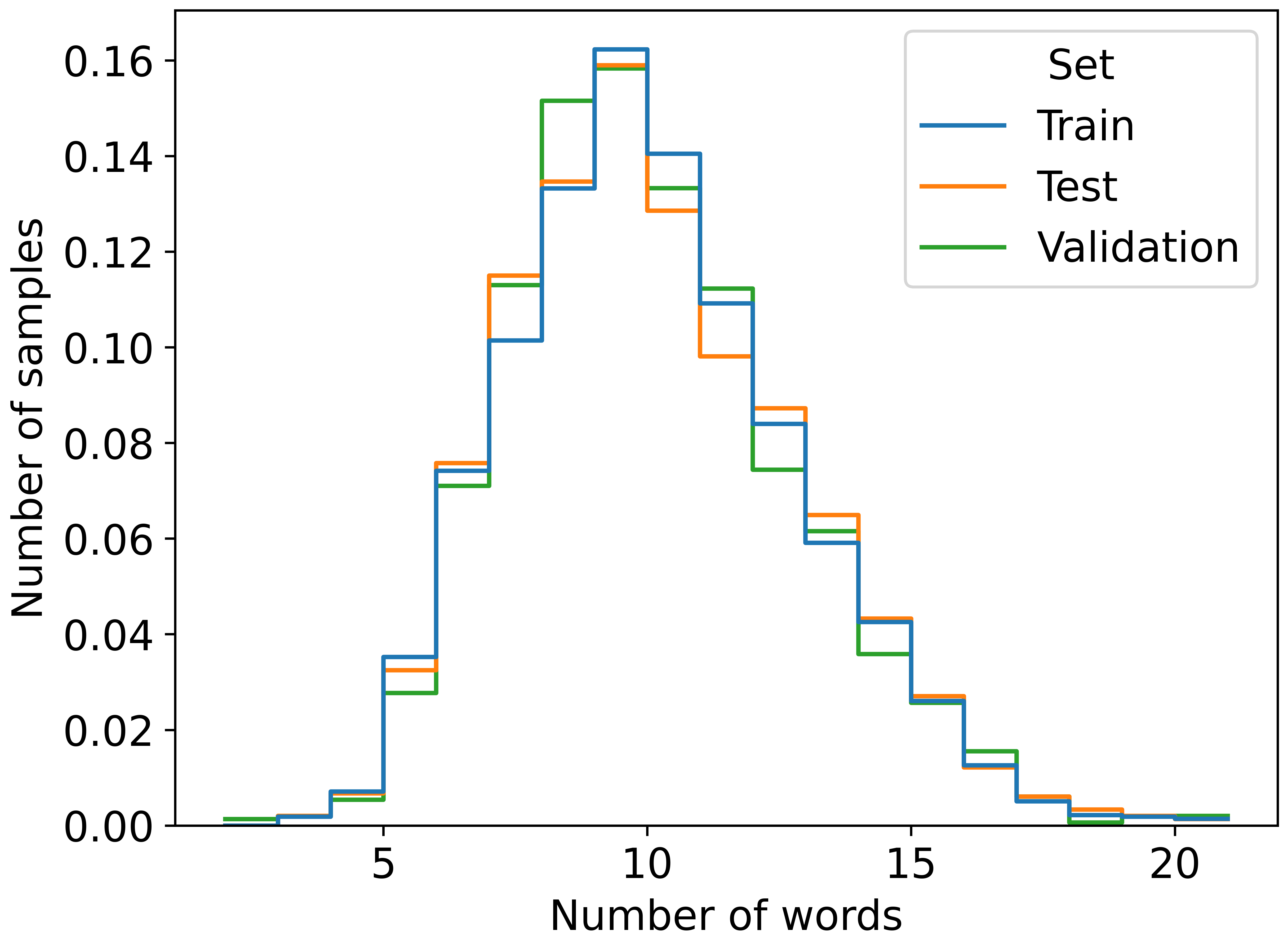}
    \caption{Distribution of text sample length, expressed in terms of word count, for the training, validation and test sets of \texttt{LETZ-SYN}.}
    \label{fig:sample_length}
\end{figure}

Figure \ref{fig:sample_length} shows the distribution of the sample length of \texttt{LETZ-SYN}, expressed as word count, and Table \ref{tab:examples} shows a small example subset of \texttt{LETZ-SYN}. Both datasets, \texttt{LETZ-SYN} and \texttt{LETZ-WoT}, are publicly available under a \textit{Creative Commons Attribution 4.0 International} (CC BY 4.0) license.

\input{Tables/examples}

\section{Implementation Details} \label{app: implementation}

\subsection{Methodology}
\begin{figure}[h!]
    \centering
    \begin{subfigure}[b]{\linewidth}
       \includegraphics[width=\linewidth]{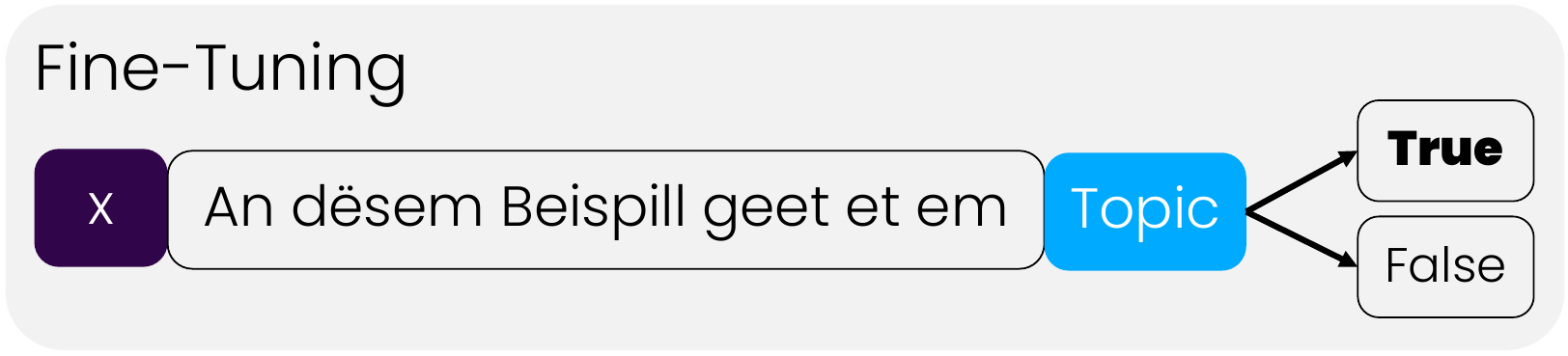}
       \caption{The model is fine-tuned on detecting whether a topic is present in a sample \textbf{x} or not (= binary classifier). \underline{Translation}: \textit{This example is about...}}
       \label{fig:fine-tuning} 
    \end{subfigure}

     \vspace{10pt}
     
    \begin{subfigure}[b]{\linewidth}
       \includegraphics[width=\linewidth]{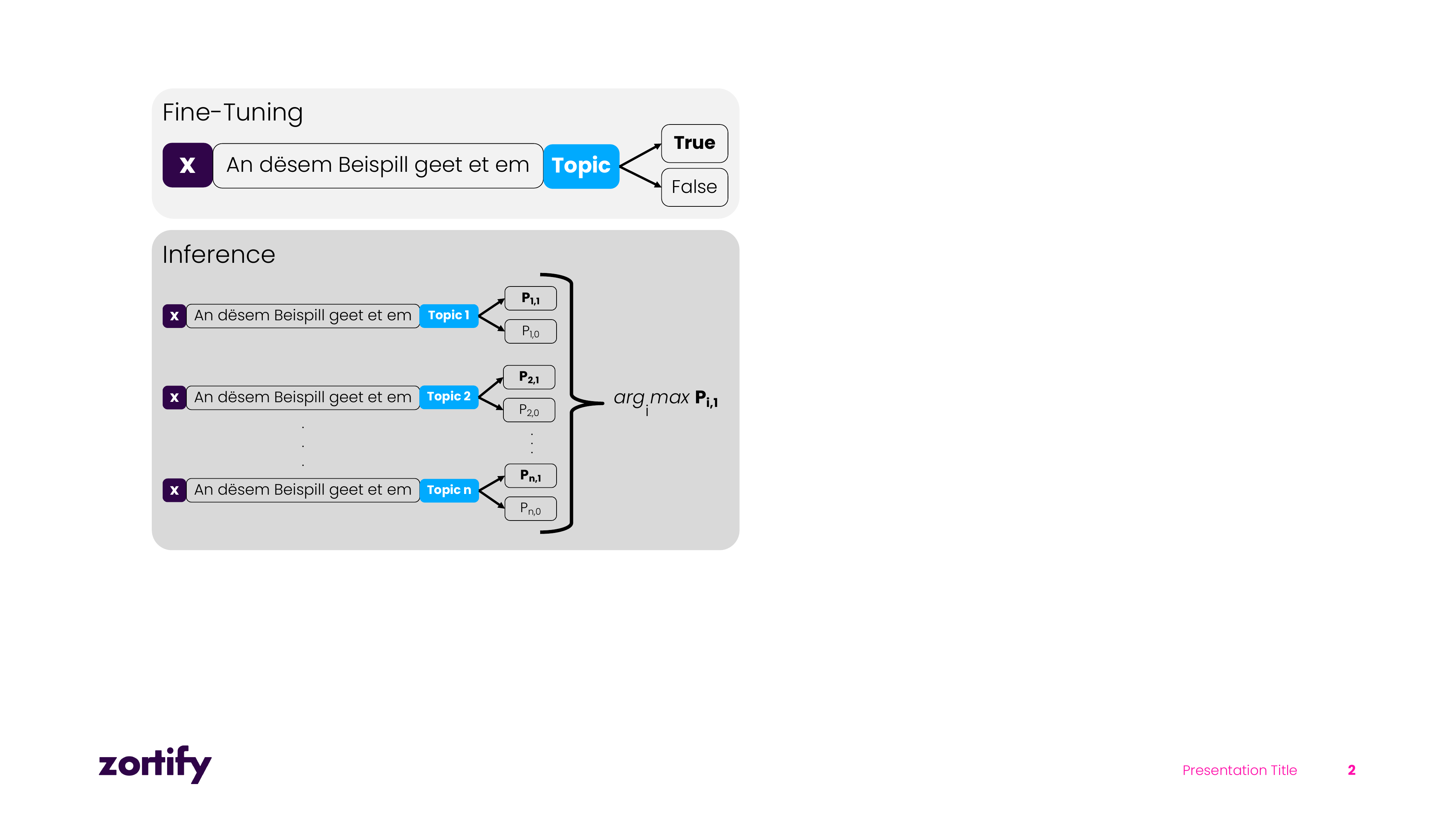}
       \caption{The model estimates the likelihood of each candidate topic independently at the inference stage and then the topic with the maximum probability is chosen.}
       \label{fig:evaluation}
    \end{subfigure}
    
    \caption{Illustration of the \textit{entailment approach} \citep{yin_benchmarking_2019} for ZSC.}
    \label{fig:entailment_approach}
    
\end{figure}

We provide a visual illustration of the \textit{entailment approach} \citep{yin_benchmarking_2019} that we use in our experiments in Figure \ref{fig:entailment_approach}. The natural language label description words and number of samples per class during evaluation are provided in Table \ref{tab:evaluation_sets}.

\subsection{Models}
We conduct our experiments on the base multilingual BERT (cased) \citep{devlin_bert_2019} and LuxemBERT \citep{lothritz_luxembert_2022} models. Both models are based on the same architecture and have 12 attention heads and
12 transformer blocks with a hidden size of 768. 
mBERT and LuxemBERT have a vocabulary size of 30,000 and 119,547 respectively. Both models have 110 million parameters.

\subsection{Reproducibility}
To reduce the computational expenses, we refrain from conducting hyper-parameter tuning and employ the configurations that yielded satisfactory results in our initial experiments. We conduct all the experiments using the AdamW optimizer \citep{loshchilov_decoupled_2019} with a learning rate of 2e-5 with 10\% warm-sup steps and linear decay and a batch size of 32. We fine-tune, with 10 warm-up steps, over 5 epochs. We perform validation after each epoch and select the optimal checkpoint based on the lowest validation loss. The maximum sequence length, during training, is set to 128 tokens. During evaluation, we set the maximum length to 128 tokens for SIB-200, and to 512 for the LuxNews dataset. For each evaluation dataset, we output the accuracy and macro-averaged F1 score.

\subsection{Computational Resources}
All experiments were run within a few hours on 4 A100 40GB GPUs in parallel, using 4 different random seeds (one per GPU).

\input{Tables/eval_datasets}

%% file: Tables/examples.tex
\begin{table*}[b]
    \centering
    \setlength{\tabcolsep}{10pt} 
    \renewcommand{\arraystretch}{1.5} 
    \begin{tabular}{llc}
        \hline
        Text & Label & Class \\
        \hline \hline
        Gedëlleg dech a waart op de richtegen \textbf{Abléck}! & Moment & 1 \\
        (\textit{Be patient and wait for the right \textbf{point in time}!}) & (\textit{moment}) & \\
        \hline
        Däin Auto huet hannen um Parechoc eng Téitsch. & Libell & 0 \\
        (\textit{Your car has a dent on the rear bumper.}) & (\textit{dragon-fly}) & \\
        \hline
        Bei esou vill Kandidate muss eng \textbf{Auswiel} gemaach ginn. & Selektioun & 1 \\
        (\textit{With so many candidates, a \textbf{choice} must be made.}) & (\textit{selection}) & \\
        \hline
        Ech schécken der d'Adress vun engem lëschtege Site.	& Schrauwenzéier & 0 \\
        (\textit{I am sending you the link to a funny website.}) & (\textit{screwdriver}) & \\
        \hline \hline
    \end{tabular}
    \caption{Examples from our dataset (\textit{with English translations}).}
    \label{tab:examples}
\end{table*}

%% file: Tables/eval_datasets.tex
\begin{table*}[b]
\centering
\renewcommand{\arraystretch}{1.5}
\setlength{\tabcolsep}{20pt}

\begin{tabular}{lllcc}
\hline
\textbf{Dataset} & \textbf{Class} & \textbf{Class Label} & \textbf{n}\\ \hline

\multirow{8}{*}{LuxNews} & \textbf{Sports} & \texttt{Sport} & 567 \\
 & \textbf{Culture} & \texttt{Konscht} & 266 \\
 & \textbf{Technology} & \texttt{Technologie} & 199 \\
 & \textbf{Gaming} & \texttt{Videospiller} &  82 \\
 & \textbf{Cooking recipes} & \texttt{Rezept} & 20 \\ 
 & National news & / & \\
 & International news & / & \\
 & European news & / & \\ \hline

\multirow{7}{*}{SIB-200} & \textbf{Science/Technology} & \texttt{Technologie} & 51 \\
 & \textbf{Travel} & \texttt{Rees} & 40 \\
 & \textbf{Politics} & \texttt{Politik} & 30 \\
 & \textbf{Sports} & \texttt{Sport} & 25 \\
 & \textbf{Health} & \texttt{Gesondheet} & 22 \\ 
 & \textbf{Entertainment} & \texttt{Entertainment} & 19 \\ 
 & \textbf{Geography} & \texttt{Geografie} & 17 \\ \hline

\end{tabular}
\caption{The original classes and their corresponding translated Luxembourgish class labels that were used our experimental setup. We used the classes marked in \textbf{bold} for evaluation, and discarded the rest from the evaluation set. \textbf{n} is the number of samples used for evaluation.}
\label{tab:evaluation_sets}
\end{table*}